\documentclass{article}

\usepackage{arxiv}

\usepackage{amsmath,amssymb,amsfonts}
\usepackage{algorithmic}
\usepackage{graphicx}
\usepackage{textcomp}
\usepackage{comment}
\usepackage{booktabs}       
\usepackage{xcolor}
\usepackage[backend=bibtex, isbn=false, url=false, sorting=none]{biblatex}
\def\BibTeX{{\rm B\kern-.05em{\sc i\kern-.025em b}\kern-.08em
    T\kern-.1667em\lower.7ex\hbox{E}\kern-.125emX}}

\usepackage[utf8]{inputenc} 
\usepackage[T1]{fontenc}    
\usepackage{hyperref}       
\usepackage{url}            
\usepackage{amsfonts}       
\usepackage{nicefrac}       
\usepackage{microtype}      
\usepackage{lipsum}		
\usepackage{doi}

\title{Lightweight Neural Architecture Search for Cerebral Palsy Detection}

\newif\ifuniqueAffiliation
\uniqueAffiliationtrue

\ifuniqueAffiliation 
\author{ \href{https://orcid.org/0009-0005-6310-408X}{\includegraphics[scale=0.06]{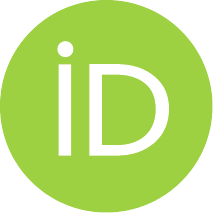}\hspace{1mm}Felix Tempel}\\
	Faculty of Informatics\\
	Norwegian University of Science and Technology\\
	Trondheim, Norway \\
	\texttt{felix.e.f.tempel@ntnu.no} \\
    \And
    \href{https://orcid.org/0000-0002-2469-1809}{\includegraphics[scale=0.06]{orcid.pdf}\hspace{1mm}Espen Alexander F. Ihlen} \\
	Faculty of Medicine and Health Sciences\\
	Norwegian University of Science and Technology\\
	Trondheim, Norway \\
	\texttt{espen.ihlen@ntnu.no} \\
	\AND
	\href{https://orcid.org/0000-0003-1820-6544}{\includegraphics[scale=0.06]{orcid.pdf}\hspace{1mm}Inga Strümke} \\
	Faculty of Informatics\\
	Norwegian University of Science and Technology\\
	Trondheim, Norway \\
	\texttt{inga.strumke@ntnu.no} \\
}

\begin{document}

\maketitle
\begin{abstract}
The neurological condition known as cerebral palsy (CP) first manifests in infancy or early childhood and has a lifelong impact on motor coordination and body movement. 
CP is one of the leading causes of childhood disabilities, and early detection is crucial for providing appropriate treatment.
However, such detection relies on assessments by human experts trained in methods like general movement assessment (GMA). 
These are not widely accessible, especially in developing countries.
Conventional machine learning approaches offer limited predictive performance on CP detection tasks, and the approaches developed by the few available domain experts are generally dataset-specific, restricting their applicability beyond the context for which these were created.
To address these challenges, we propose a neural architecture search (NAS) algorithm applying a reinforcement learning update scheme capable of efficiently optimizing for the best architectural and hyperparameter combination to discover the most suitable neural network configuration for detecting CP.
Our method performs better on a real-world CP dataset than other approaches in the field, which rely on large ensembles.
As our approach is less resource-demanding and performs better, it is particularly suitable for implementation in resource-constrained settings, including rural or developing areas with limited access to medical experts and the required diagnostic tools.
The resulting model's lightweight architecture and efficient computation time allow for deployment on devices with limited processing power, reducing the need for expensive infrastructure, and can, therefore, be integrated into clinical workflows to provide timely and accurate support for early CP diagnosis.
Finally, our proposed lightweight approach simplifies the application of computationally intensive explainability methods, as it avoids the scenario where a full ensemble must be explained. The code is publicly available at https://github.com/DeepInMotion/AutoCP.
\end{abstract}


\section{Introduction}
Cerebral palsy (CP) is one of the most common childhood disabilities, characterized by physical and functional limitations caused by developmental brain damage \cite{patelCerebralPalsyChildren2020, rosenbaumReportDefinitionClassification2007}.
The diagnosis usually happens 12-24 months after labor and is best supplemented with a follow-up treatment to guide the needed interventions for the infant \cite{morganEarlyInterventionChildren2021, maitreImplementationHammersmithInfant2016}.
The general movement assessment (GMA) is considered the gold standard in diagnosing CP \cite{prechtlEarlyMarkerNeurological1997}. 
However, providing physicians with adequate GMA training takes time and resources and necessitates manual examination of the infant's movements. 
In addition, the procedure is vulnerable to subjective evaluation and dependent on the physician's expertise \cite{zhangCerebralPalsyPrediction2022}.

While the traditional GMA method involves manual observation and analysis, recent advancements in machine learning have enabled the utilization of automated methods. 
This integration aims to streamline the process, reduce subjectivity, and potentially make GMA more accessible and efficient \cite{silvaFutureGeneralMovement2021, marcroftMovementRecognitionTechnology2015}.
Some early efforts on automating the GMA for CP detection are proposed in \cite{orlandiDetectionAtypicalTypical2018, stahlOpticalFlowBasedMethod2012, ihlenMachineLearningInfant2019}.
Despite their promising results, these methods primarily depend on handcrafted features and traditional algorithms, which limit their ability to capture the full complexity of infant CP-related movements. 
Moreover, manual labeling induces bias and limits the system's adaptability because these models are not well-suited to generalize outside their distinct training contexts and predetermined parameters.

Neural Architecture Search (NAS) has significantly enhanced the development of machine learning models, making the process more efficient and reducing the need for manual intervention \cite{renComprehensiveSurveyNeural2022}. 
By automating the design of neural networks, NAS optimizes architectures tailored to specific tasks and datasets, minimizing the need for extensive domain expertise and thus making it more accessible to a larger audience \cite{whiteNeuralArchitectureSearch2023}.
With the use of NAS, it is possible to overcome limitations inherent in traditional GMA automation methods, often constrained by domain-specific designs and data dependencies.
NAS hereby enhances the flexibility of the obtained models and significantly improves their scalability and adaptability across diverse clinical environments and datasets, making them more robust and democratizing their usability.

An important contribution in this domain is the work of \cite{groosDevelopmentValidationDeep2022}, where the authors employ a deep learning-based method with graph convolutional networks on infant skeleton data to detect CP.
However, their results rely on a resource-intensive ensemble prediction strategy, requiring the training and evaluation of 70 individual models.
This significantly limits the applicability in resource-constrained settings and causes post-hoc explanation methods to be impractical to apply.

We utilize a NAS algorithm with a refined architecture and hyperparameter search space on a real-world CP skeleton dataset to address this issue, constructing a lightweight architecture.
Our approach yields a smaller architecture that outperforms existing methods, including a large ensemble, in terms of both Sensitivity and resource efficiency. 
As a result, our proposed NAS is a promising tool for practitioners with limited machine learning experience and limited availability of computational resources, helping to further democratize and improve the accessibility of CP detection.
Our contributions can be summarized as follows:
\begin{itemize}
    \item We employ a NAS algorithm with an expanded search space and tailored building blocks for the CP skeleton dataset.
    \item We present a final lightweight architecture that outperforms a large ensemble and other methods while also being more resource-efficient. 
    \item Democratization of automated CP prediction.
\end{itemize}

\section{Method}
The NAS algorithm aims to optimize hyperparameters, denoted as $h$, and the architecture components, denoted as $\alpha$, building upon methods introduced in \cite{dongAutoHASEfficientHyperparameter2021, tempelAutoGCNgenericHumanActivity2024}.
The search space for the NAS is defined as $\mathcal{S} \in \{\alpha, h\}$, from which the reinforcement controller can sample.
Table \ref{tab:sp_cp} outlines the parameters and their respective value ranges.
During the NAS process, the loss function $\mathcal{L}$ on the validation set $\mathcal{D}_{val}$ is minimized while guaranteeing optimal parameters, $\omega_{\alpha, h}^*$, that are obtained from the minimized loss on both the architecture, $\alpha$, hyperparameters, $h$, and the training dataset, $\mathcal{D}_{train}$:
\begin{equation}
    \min_{\alpha, h} \mathcal{L}(\alpha, \omega_{\alpha, h}^*, \mathcal{D}_{val})
    \quad \text{s.t.} \quad \omega_{\alpha, h}^* = \mathcal{L}(\alpha, h, \mathcal{D}_{train}).
\end{equation}

\subsection{Infant Skeleton}
\label{sec:infant}
Several features are computed from the infant skeleton and used as inputs for the NAS.
These features are grouped into four categories: 1) position ($P$), 2) velocity ($V$), 3) bone ($B$), and 4) acceleration ($A$). 
The input skeleton sequence is $\mathcal{X} \in \mathcal{R}^{C, T, V}$ where $C$ denotes the $x$, and $y$ coordinates, $T$ represents the number of frames, and $V$ specifies the vertex, i.e.,\ joint.

\begin{figure}[t]
    \centering
    \includegraphics[scale=2.5]{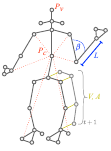}
    \caption{CP infant skeleton with 29 body key points and the respective input features $P \in \{P_v, P_c\}$, $V$, $A$, and $B \in \{L, \beta\}$.}
    \label{fig:cp_skeleton}
\end{figure}

The processing of $\mathcal{X}$ includes extracting the joint positions ($P_v$) and those relative to the central joint ($P_c$), velocities, and acceleration with the latter two estimated using finite differences. 
An 8th-order Butterworth filter is applied to the acceleration data to enhance signal quality, while bone features are calculated based on joint distances and angles.
The skeletons used in the training process are randomly augmented through scaling translation and rotation to prevent overfitting. 
Fig.~\ref{fig:cp_skeleton} illustrates the infant skeleton and the feature categories.

\subsection{Architecture}
\begin{figure*}[]
    \centering
    \includegraphics[scale=1]{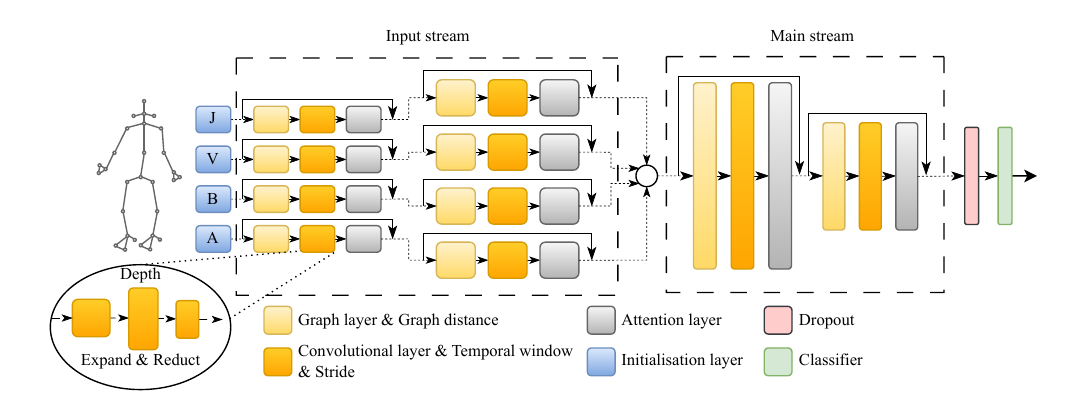}
    \caption{Architecture building blocks for the input- and main stream search block.}
    \label{fig:cp_archi}
\end{figure*}

An overview of the architecture and the respective searchable streams from $\mathcal{S}$ is shown in Fig.~\ref{fig:cp_archi}.
Each input feature is processed through a distinct input stream, and the outputs of these streams are subsequently fused. 
This output is passed through the main stream before being fed into the classifier.

The convolutional layers (emphasized in orange) are flexible, allowing for depth, stride, internal expansion, or reduction variations depending on the respective search parameters.
The input stream consists of repetitive building blocks that can be scaled through the \textit{expand ratio}. 
Conversely, the main stream undergoes a reduction in layer sizes, controlled by the \textit{reduction ratio} parameter.

\section{Experiments}
\subsection{Dataset}
The dataset used contains 557 videos of infants with medical risk factors for CP, recorded between 2001 and 2018 across multiple countries: the United States \((n=248)\), Norway \((n=190)\), Belgium \((n=37)\), and India \((n=82)\).
Two experts have classified these recordings of 9 to 18 weeks post-term infants according to the standardized criteria in \cite{prechtlEarlyMarkerNeurological1997}.
For each frame in a video, the positions of 29 body key points of an infant in motion are represented using $x$ and $y$ coordinates, forming the infant skeletons as shown in Fig.~\ref{fig:cp_skeleton}.
The skeleton sequence is preprocessed, which includes resampling at 30 Hz, temporal smoothing via a 5-point median filter, and coordinate standardization relative to the trunk length.
This standardization involves centering around the median mid-pelvis position and normalizing by twice the infant's trunk length.
The data is segmented into windows lasting 5 seconds each, with a 2.5-second overlap between consecutive windows.
Finally, the input features are computed as outlined in Section \ref{sec:infant}.
For the NAS procedure, the dataset is split into three subsets: a training set, $\mathcal{D}_{\text{train}}$, and a validation set, $\mathcal{D}_{\text{val}}$, from which $n=63$ are CP and $n=355$ are without CP related movements. 
The third subset, $\mathcal{D}_{\text{test}}$, contains $n=139$ samples, of which $n=21$ display CP-related movements and $n=118$ show no CP-related movements.

\subsection{Implementation}
A student architecture configuration is randomly sampled from the search space $\mathcal{S}$ and trained for $50$ epochs.
The training employs an initial warm-up phase of $10$ epochs with a linearly increasing learning rate up to the sampled value from $\mathcal{S}_{h}$.
Furthermore, an early stopping criterion is utilized at the $6th$ epoch if the student architecture Area Under the Curve (AUC) is below $0.5$.
Within each controller update iteration, $30$ student architectures are trained.  
Subsequently, the controller updates its internal state, where the state values represent the probabilities of selecting components from $\mathcal{S}$. 
These probabilities are adjusted based on the reward $r$ derived from the AUC achieved by the student architectures.
The controller employs a learning rate of $0.001$ and is optimized with the Adam optimizer.
If the trained student architectures achieve an AUC of $0.9$ or higher, these are saved in the replay memory, which serves as a repository of high-performing models for reuse in subsequent controller update cycles.
Following this, a candidate with the highest state values for $\mathcal{S}_{h}$ and $\mathcal{S}_{a}$ is built and trained for an extended period of 300 epochs.
The learning rate is halved at 200 and 250 epochs to facilitate convergence.

The experiments are performed on a single NVIDIA-RTX 3090 with 24 GB GPU RAM on the PyTorch framework (version 2.3.1) \cite{paszkePyTorchImperativeStyle2019}. 
A global seed of $1234$ is set for reproducibility.

\begin{table}[]
\centering
\caption{The search space for the CP dataset, comprising parameter and value ranges, grouped into effective areas. The search space includes hyperparameters $h$ for the optimizer and architectural choices $\alpha$.}
\label{tab:sp_cp}

    \begin{tabular}{@{}lll@{}}
        \toprule
        \textbf{Parameter}        & \textbf{Possible choices}                                       & \textbf{Best choice} \\ \midrule
        \multicolumn{3}{l}{\textbf{General}}                                             \\ \midrule
        Init layer size         & {[}16, 32, 48, 64, 96{]}                                      & 64 \\
        Activation layer        & {[}Relu, Relu6, Hardswish, Swish{]}                           & Swish \\
        Attention layer         & {[}Stja, Ca, Fa, Ja, Pa{]}                                    & Fa \\
        Conv. layer type        & {[}Basic, Bottleneck, Sep, SG, V3, Shuffle{]}                 & Sep     \\
        Dropout probability     & {[}0, 0.025, 0.05, 0.1{]}                                     & 0.05 \\ 
        Multi-GCN               & {[}True, False{]}                                             & False \\ 
        Expand ratio            & {[}1, 1.5, 2{]}                                               & 1.5 \\ 
        Reduction ratio         & {[}1, 1.5, 2{]}                                               & 1.5 \\ \midrule
        \multicolumn{3}{l}{\textbf{Input stream}}                                             \\ \midrule
        Blocks input               & {[}1, 2, 3{]}                                              & 2 \\
        Depth input                & {[}1, 2, 3{]}                                              & 2 \\
        Stride input               & {[}1, 2, 3{]}                                              & 3 \\
        Scale input                & {[}0.8, 0.9, 1, 1.1, 1.2{]}								& 1 \\
        Temporal window input      & {[}3, 5, 7{]}                                              & 3 \\
        Graph distance input    & {[}1, 2, 3{]}                                                 & 1 \\ \midrule
        \multicolumn{3}{l}{\textbf{Main stream}}                                              \\ \midrule
        Blocks main             & {[}1, 2, 3, 4{]}                                              & 2 \\
        Depth main              & {[}1, 2, 3, 4{]}               							    & 2 \\
        Stride main             & {[}1, 2, 3{]}                                                 & 1 \\
        Scale main              & {[}0.95, 1, 1.1, 1.2, 1.3{]}                                  & 1 \\
        Temporal window main    & {[}3, 5, 7{]}                                                 & 7 \\
        Graph distance main     & {[}1, 2, 3{]}                                                 & 1 \\ \midrule
        \multicolumn{3}{l}{\textbf{Optimizer}}                                                \\ \midrule
        Optimizer               & {[}SGD, Adam, AdamW{]}                                        & Adam \\ 
        Learning rate           & {[}0.005, 0.001, 0.0005{]}                                    & 0.005 \\
        Weight decay            & {[}0.0, 0.01, 0.001, 0.0001{]}							    & 0.0 \\
        Momentum                & {[}0.5, 0.9, 0.99{]}									        & 0.99     \\
        Batch size              & {[}24, 32, 40{]}											    & 32 \\ \bottomrule
    \end{tabular}
\end{table}

\subsection{Comparison with other methods}
In Table \ref{tab:results}, a performance comparison of various methods on the external test set $\mathcal{D}_{test}$ is shown, including our NAS method, the large ensemble from \cite{groosDevelopmentValidationDeep2022}, the GMA method \cite{prechtlEarlyMarkerNeurological1997}, and the more conventional approach from \cite{ihlenMachineLearningInfant2019}. 

Our method exhibits a Sensitivity of 76.2\% and a Specificity of 93.2\%, reflecting strong negative case identification and a reasonable positive case detection.
In comparison, the large ensemble from \cite{groosDevelopmentValidationDeep2022} achieves a lower Sensitivity of 71.4\% but a slightly higher Specificity of 94.1\%.
The GMA method \cite{prechtlEarlyMarkerNeurological1997} shows comparable results, with a Sensitivity of 70.0\% and Specificity of 88.7\%.
The conventional approach from \cite{ihlenMachineLearningInfant2019} shows inferior performance compared to our NAS on all metrics.
Notably, our method produces fewer false negatives (FN = 5) than the other methods (FN = 6). 
Reducing false negative cases is particularly important in a clinical setting, as a delayed or even missed diagnosis can impede the necessary treatment. 
In contrast, false positives may lead to strain on healthcare resources and put unnecessary stress on the infants' parents. 
Still, a lower false negative count is preferable in a clinical setting, which indicates that our method holds promise in comparison with the other approaches -- although this is, given the small statistical sample, only an indication.

\begin{table*}
    \centering
    \caption{Performance comparison of different methods on the external test set $\mathcal{D}_{test}$. 
    The metrics include true positives (TP), false positives (FP), true negatives (TN), false negatives (FN), Sensitivity, Specificity, and Accuracy. 
    Furthermore, the 95\% confidence intervals calculated with the Clopper-Pearson method are provided.}
    \label{tab:results}
    \begin{tabular}{cccccccc}
        \toprule
        Method & TP & FP & TN & FN & Sensitivity & Specificity & Accuracy\\ 
        \midrule
        NAS & 16 & 8 & 110 & 5 & \textbf{76.2} (52.8-91.8)& 93.2 (87.1-97.0) & \textbf{90.6} (84.5-94.9) \\
        Ensemble \cite{groosDevelopmentValidationDeep2022} & 15 & 7 & 111 & 6 & 71.4 (47.8-88.7) & \textbf{94.1} (88.2-97.6) & \textbf{90.6} (84.5-94.9) \\
        GMA \cite{prechtlEarlyMarkerNeurological1997} & 14 & 13 & 102 & 6 & 70.0 (45.7-88.1) & 88.7 (81.5-93.8) & 85.9 (78.9-91.3) \\

        Conventional \cite{ihlenMachineLearningInfant2019} & 15 & 32 & 86 &  6 & 71.4 (47.8-88.7) & 72.9 (63.9-80.7) & 72.7 (64.5-79.9) \\
        
         \bottomrule
    \end{tabular}
    
\end{table*}

\subsection{Architectural Performance Analysis}
A comparison of the number of parameters and MACs of our proposed NAS to the ensemble \cite{groosDevelopmentValidationDeep2022} is shown in Table \ref{tab:performance}.
Our method outperforms the ensemble in terms of both of these values. 
While the full ensemble of models requires 22.883 million parameters and 126.567 billion MACs, our approach utilizes only 0.621 million parameters and 0.909 billion MACs.

This showcases the efficacy of our NAS method in developing a compact architecture that significantly enhances computational efficiency without compromising performance, making it suitable for resource-constrained applications.

\begin{table}
    \centering
    \caption{Comparison of model performance based on parameters and MACs, with parameters expressed in millions (M) and MACs in billions (G). 
    Ten ensembles are trained with the bagging strategy on seven subsets of $\mathcal{D}_{train}$ and $\mathcal{D}_{val}$, resulting in 70 individually trained models.}
    \label{tab:performance}
    \begin{tabular}{ccc}
        \toprule
        Model & Params (M) & MACs (G)\\
        \midrule
        1  & 1.099 & 11.102 \\
        2 & 0.385 & 4.109 \\
        3 & 0.882 & 8.365 \\
        4 & 2.254 & 23.233 \\
        5 & 0.546 & 4.893 \\
        6 & 1.225 & 6.853 \\
        7 & 9.653 & 35.112 \\
        8 & 1.750 & 8.827 \\
        9 & 3.199 & 9.219 \\
        10 & 2.989 & 25.956 \\
        \midrule
        Full ensemble & \textbf{22.883} & \textbf{126.567}\\
        \midrule
        NAS & \textbf{0.621} & \textbf{0.909}\\
        \bottomrule
    \end{tabular}
\end{table}

\section{Discussion}
One of the notable advantages of our approach is the efficiency in arriving at an optimal architecture. 
Compared to the large ensemble method, our NAS significantly reduces the required time without compromising the performance while also achieving a higher sensitivity.

When evaluating the performance of our model, it is important to consider the balance between sensitivity and specificity. 
The thresholds for these metrics can significantly influence the model's ability to classify true positives and negatives correctly.
In clinical applications, particularly in CP detection, optimizing this balance to minimize false negatives, which could lead to missed diagnoses, and false positives, which could cause unnecessary interventions, has to be decided.

In the healthcare domain, the aspect of explainability is particularly important since any decision made by a medical practitioner comes with a risk to a human life \cite{chaddadSurveyExplainableAI2023}.
When diagnosing an infant with CP, the physician first performs an examination and carefully concludes before going on to form and give an explanation of the diagnosis based on the symptoms and examination \cite{patelCerebralPalsyChildren2020}.
While our proposed NAS can be a useful decision support tool for physicians to use in the diagnosing process, it must still be explainable in order to become trustworthy and gain acceptance among healthcare professionals. Healthcare providers and decision-makers must be able to assess the reliability of models used in real-world clinical settings by understanding the underlying reasoning behind their predictions. 
Therefore, any model used in such a clinical setting would have to be made understandable to the end-users, usually by means of explainable AI (XAI) methods \cite{lohApplicationExplainableArtificial2022}.
In this regard, our proposed architecture is greatly beneficial, as it is easier to explain a single model than a large ensemble, both from a methodological and a computational perspective.

The limitations of our approach lie mainly in the data availability:
The used dataset holds different subtypes of CP, including spastic bilateral and spastic unilateral CP, and also different severities of these \cite{groosDevelopmentValidationDeep2022}. 
Our current approach does not distinguish between the different subtypes of CP or their severity. Furthermore, the representation of CP cases is limited relative to the number of healthy infants in the dataset.
A larger dataset with a broader range and more positive CP cases could improve the model's ability to generalize and achieve a higher Sensitivity. 
Although collecting this kind of sensitive medical data is challenging, especially as only 1.5–3 per 1000 live births develop CP \cite{patelCerebralPalsyChildren2020, morganEarlyInterventionChildren2021, johnsonCerebralPalsiesEpidemiology2000}, future research should aim at including a larger and more diverse dataset that includes a wider range and higher amount of CP cases, encompassing different subtypes and severities, to improve the resulting model's generalizability and Sensitivity.

\section{Conclusion}
This study demonstrates the effectiveness of employing a NAS algorithm on a real-world CP dataset, achieving performance metrics better than GMA, conventional machine learning and a large ensemble method. 
Our findings underscore the ability of our NAS to offer a higher Sensitivity while significantly reducing the complexity compared to other methods.
Future research should focus on further refining the NAS algorithm, incorporating larger and more diverse datasets, and providing explainability of the discovered models to enhance their clinical applicability.

\printbibliography

\end{document}